\documentclass[runningheads]{llncs}
\usepackage[T1]{fontenc}
\usepackage{graphicx,verbatim}
\usepackage{microtype}
\usepackage{hyperref}
\usepackage{color}

\usepackage{booktabs}
\usepackage{multirow}
\usepackage{amsmath}
\usepackage{amsfonts}
\usepackage{subcaption}
\usepackage[table]{xcolor}
\usepackage{colortbl}
\usepackage{tabularx}
\usepackage{array}

\definecolor{clscolor}{HTML}{FFF2CC}   
\definecolor{mlcolor}{HTML}{DAE8FC}    
\definecolor{detcolor}{HTML}{D5E8D4}   
\definecolor{cntcolor}{HTML}{FFE6CC}   
\definecolor{rgcolor}{HTML}{E1D5E7}    

\newcolumntype{Y}{>{\centering\arraybackslash}X}

\begin{document}
\title{An Empirical Analysis of Continual Learning for Heterogeneous Medical Visual Question Answering}
\titlerunning{An Empirical Analysis of Continual Learning for Heterogeneous MedVQA}
%

\author{Mai A. Shaaban$^{*,1,2}$ \and Tausifa Jan Saleem$^{*,1}$ \and Alaa Mohamed$^{1}$ \and Dilnaz Utemissova$^{1}$ \and Ufaq Khan$^{1}$ \and Mohammad Yaqub$^{1}$}
\authorrunning{Shaaban et al.}
\institute{
    $^{1}$ Mohamed bin Zayed University of Artificial Intelligence, Abu Dhabi, UAE \\
    $^{2}$ Department of Mathematics and Computer Science, Faculty of Science, Alexandria University, Alexandria, Egypt \\
    \email{\{mai.kassem, tausifa.saleem, Alaa.Mohamed, Dilnaz.Utemissova, Ufaq.Khan, mohammad.yaqub\}@mbzuai.ac.ae}
}

\footnotetext{* These authors contributed equally to this work.}
  
\maketitle              
\begin{abstract}

Deploying medical visual question answering (MedVQA) systems in real-world clinical settings requires models that adapt to new clinical tasks without forgetting previously acquired knowledge. Continual learning (CL) provides a practical framework for this setting. Despite rapid progress in medical vision-language models, the behavior of CL methods when training these models across heterogeneous MedVQA tasks remains underexplored. This work presents a systematic evaluation of CL for MedVQA across diverse clinical objectives, including classification, multi-label classification, detection, cell counting, and report generation. Specifically, we explore (1) the ability of existing CL methods to mitigate catastrophic forgetting; (2) their sensitivity to task ordering, analyzing how different task sequences influence performance retention and forgetting; and (3) the evolution of low-rank adaptation parameters as new tasks are learned, revealing patterns of weight drift under different CL methods. Our findings suggest that existing CL methods struggle to maintain stability-plasticity balance when tasks with different objectives and supervision formats are interleaved. Code and full experimental setup will be publicly available.


\end{abstract}
\section{Introduction}
\begin{figure}[t!]
    \centering
    \includegraphics[width=0.99\linewidth]{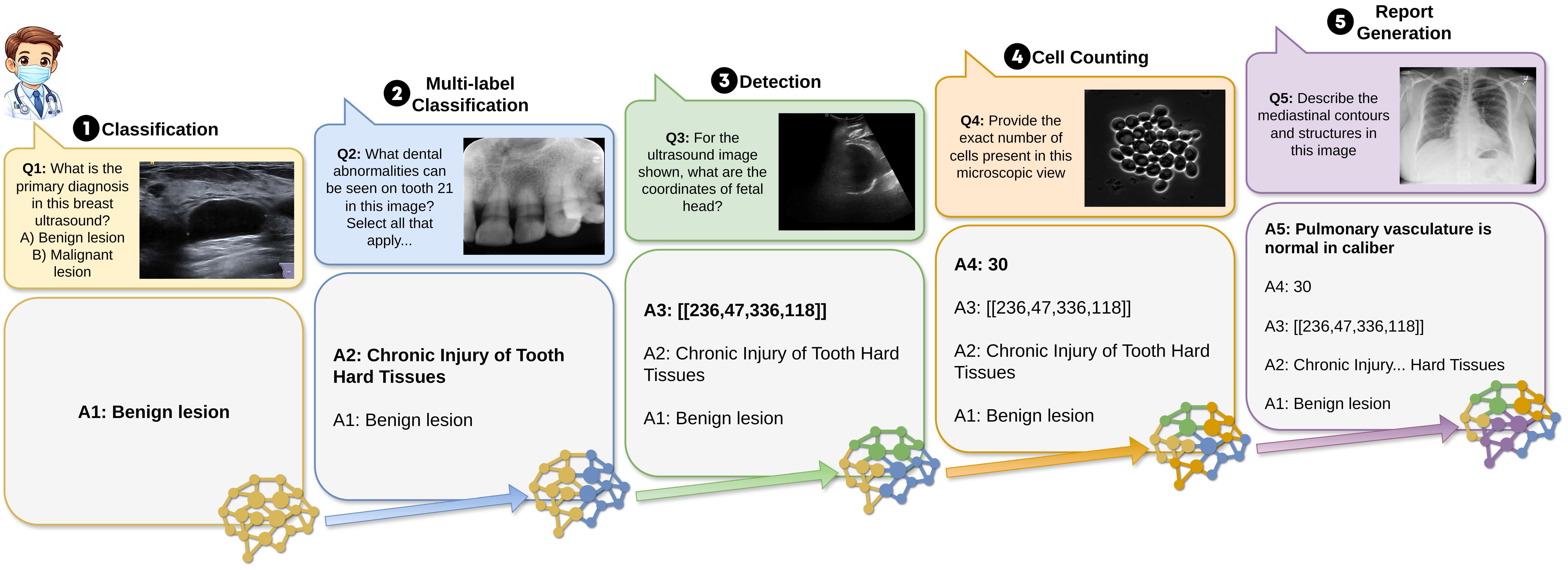} 
    \caption{CL for heterogeneous MedVQA tasks. An illustration of the desired stability-plasticity behavior, where the model is progressively updated to incorporate new tasks while preserving previously acquired knowledge.}
    \label{fig:motivational}
\end{figure}
Medical visual question answering (MedVQA) has emerged as a powerful paradigm for unified reasoning over medical images and natural language, enabling a wide range of clinical tasks such as disease classification, abnormality detection, and report generation~\cite{lau2018dataset,lin2023medical,shaaban2025motor}. Recent advances in medical vision-language models (VLMs), together with parameter-efficient fine-tuning methods \cite{peft} such as low-rank adaptation (LoRA) \cite{hu2022lora}, have substantially improved performance across diverse medical datasets, positioning MedVQA systems as promising general-purpose clinical assistants \cite{kalpelbe2025vision}. Most MedVQA systems rely on joint multi-task training, assuming simultaneous access to all task data. However, in real-world clinical settings, data typically arrive incrementally and tasks evolve over time \cite{bruno2025continual,jiang2025omnidoctor}, requiring models to adapt continuously without forgetting prior knowledge.\\

Continual learning (CL) \cite{chen2018lifelong,biesialska2020continual,wang2024comprehensive,qu2025recent} provides a principled framework for this setting by enabling sequential task adaptation while achieving an effective \emph{stability-plasticity} trade-off, where the model remains sufficiently \emph{plastic} to adapt to new tasks while maintaining \emph{stability} to preserve previously acquired knowledge and prevent catastrophic forgetting \cite{chen2022continual}. Although CL has been explored in medical imaging and MedVQA, prior works predominantly focus on domain-incremental settings, where tasks differ in data source or imaging modality but share the same objective and supervision format \cite{bruno2025continual,qazi2024continual,jiang2025omnidoctor}. In contrast, real-world clinical workflows require medical VLMs to operate across structurally heterogeneous tasks that differ in objective type, supervision granularity, and semantic constraints on the generated outputs (Fig.~\ref{fig:motivational}). This heterogeneity introduces distinct optimization dynamics and cross-task interference patterns that are not captured by conventional CL benchmarks, leaving a critical gap in understanding CL behavior in MedVQA. This motivates three key research questions (RQs): \textbf{RQ1.} \textit{Do current CL methods mitigate catastrophic forgetting in heterogeneous MedVQA?}, \textbf{RQ2.} \textit{How does task ordering influence performance and forgetting across different CL methods?} and \textbf{RQ3.} \textit{How does drift in LoRA weights relate to forgetting behavior across CL methods?}

To answer these questions, we conduct a systematic study of a representative set of CL strategies under sequential training across diverse MedVQA tasks, including classification, multi-label classification, detection, cell counting, and report generation. For \textbf{RQ1}, we monitor task-wise performance throughout sequential training and compute forgetting ratios that capture relative performance drops following the learning of new tasks. For \textbf{RQ2}, we evaluate the sensitivity of each CL method across three different sequences of tasks with varying complexity based on output structure, supervision granularity, and reasoning demands. For \textbf{RQ3}, we visualize sequential LoRA parameter drift and quantify the degree of angular spread, relating these weight dynamics to observed forgetting patterns.

\section{Methods}

We study MedVQA under a CL setting, where a pre-trained medical VLM is exposed to a sequence of $T$ tasks $\{\mathcal{D}_t\}_{t=1}^{T}$ ($\mathcal{D}_t = \{(x_i^{(t)}, y_i^{(t)})\}_{i=1}^{N_t}$ denotes the dataset for task $t$). The tasks are structurally heterogeneous and differ in objective and supervision format, spanning classification, multi-label classification, detection, cell counting, and report generation. Training proceeds sequentially with the objective of optimizing performance on the current task while minimizing performance degradation on previously learned tasks. Importantly, task identity is not provided at inference time; a single shared model must generalize across all tasks without explicit task indicators. The backbone VLM remains frozen throughout training, and only LoRA parameters are updated, placing the stability-plasticity trade-off entirely within the adapter space. The overall goal is to obtain a model that performs strongly across all heterogeneous tasks while effectively mitigating catastrophic forgetting.

To systematically study this setting, we adapt a representative set of CL strategies that cover rehearsal-based, parameter-regularization, knowledge-transfer, and subspace-isolation techniques, reformulating them so that their constraints and updates operate solely on LoRA parameters. We further include two reference baselines: multi-task learning (MTL), which jointly trains a shared LoRA adapter across all tasks as an upper-bound reference, and sequential fine-tuning (SEQ-FT), which updates the same adapter sequentially without forgetting mitigation. The following presents a description of the CL methods:\\
\noindent
\textbf{Replay} mitigates catastrophic forgetting by rehearsing samples from previously learned tasks during training on new tasks~\cite{bagus2021investigation}. We adopt a rehearsal-based variant with a balanced replay buffer that stores samples from prior tasks and mixes them with data from the current task during training~\cite{merlin2022practical}.\\
\noindent
\textbf{Elastic Weight Consolidation (EWC)}~\cite{kirkpatrick2017overcoming} regularizes important parameters for previous tasks. We adapt it by restricting consolidation to LoRA parameters. After training task $t$, parameter importance is estimated using the Fisher information diagonal $F_i^{(t)} \approx \mathbb{E}_{(x,y)\sim\mathcal{D}_t}\!\left[\left(\frac{\partial \mathcal{L}(x,y;\theta)}{\partial \theta_i}\right)^2\right]_{\theta=\theta_t^*}$, where $\theta_t^*$ denotes the convergent parameters after task $t$. During training on task $t+1$, deviations from $\theta_t^*$ are penalized via 
$\mathcal{L}_{\mathrm{EWC}} = \frac{\lambda_\mathrm{EWC}}{2} \sum_i F_i^{(t)} (\theta_i - \theta_{t,i}^*)^2$, yielding the objective $\mathcal{L} = \mathcal{L}_{\mathrm{task}} + \mathcal{L}_{\mathrm{EWC}}$, where $\lambda_\mathrm{EWC}$ controls the stability-plasticity trade-off.  We implement multi-task EWC, storing a separate Fisher matrix and parameter snapshot for each completed task and summing their corresponding penalties during subsequent training.\\
\noindent
\textbf{Knowledge Distillation (KD)-based Approach} transfers knowledge across tasks using a previously trained model as a reference \cite{lwf,rebuffi2017icarl,azimi2024kd}. We adapt recent KD-LoRA formulations to a CL setting by using output-level distillation as a regularizer to preserve previously acquired knowledge. In task $t$, the student model is equipped with LoRA modules and trained on the current dataset $\mathcal{D}_t$, while the teacher model corresponds to the model obtained after completing task $t\!-\!1$ and is kept frozen. Given an input $x \sim \mathcal{D}_t$, we encourage the student to match the teacher’s predictive distribution by minimizing an output-level distillation loss $\mathcal{L}_{\mathrm{KD}} = \mathrm{KL}\!\left(\sigma\!\left(\frac{z(x;\theta^{(t-1)})}{\tau}\right) \,\middle\|\, \sigma\!\left(\frac{z(x;\theta^{(t)})}{\tau}\right)\right)$, where $z(x;\theta^{(t)})$ and $z(x;\theta^{(t-1)})$ denote the logits produced by the student and teacher models, respectively, $\sigma(\cdot)$ is the softmax function, $\mathrm{KL}(\cdot\|\cdot)$ is the Kullback--Leibler divergence and $\tau$ is the distillation temperature. The overall training objective is given by $\mathcal{L} = \mathcal{L}_{\mathrm{task}} + \lambda_{\mathrm{KD}} \mathcal{L}_{\mathrm{KD}}$, where $\mathcal{L}_{\mathrm{task}}$ is the task-specific loss on $\mathcal{D}_t$ and $\lambda_{\mathrm{KD}}$ controls the stability-plasticity trade-off.\\
\noindent
\textbf{Soft-masking of Parameter-level Gradient flow (SPG)}~\cite{konishi2023parameter} softly constrains updates to parameters that are important for previous tasks. We adapt it by restricting gradient modulation to LoRA parameters. Parameter importance is estimated via gradient sensitivity and is used to modulate future updates without hard freezing. Let $\theta$ denote the trainable parameters. During task $t$, raw gradients $g_t(\theta)=\nabla_\theta \mathcal{L}_t(\theta)$ are scaled using an accumulated importance mask as $\tilde{g}_t(\theta) = g_t(\theta)\odot\bigl(1-a_{\max}^{(t)}(\theta)\bigr)$, where $a_{\max}^{(t)}(\theta)\in[0,1]$ encodes parameter importance from previous tasks, protecting the parameters of high-importance while keeping others plastic. After training task $t$, importance is computed by probing the model with data from all observed tasks. For each $k\le t$, gradients are accumulated with respect to $\theta$ to obtain the gradient tensor $G_{t,k}(\theta)$. Each gradient tensor is standardized via $S(H)=\tanh\!\left(\frac{H-\mu(H)}{\sigma(H)+\varepsilon}\right)$, where $\mu(\cdot)$ and $\sigma(\cdot)$ denote the mean and standard deviation. The task-level mask is defined as $m^{(t)}(\theta)=\max_{k\le t}\left|S\!\left(G_{t,k}(\theta)\right)\right|$, and the accumulated mask used for subsequent training is $a_{\max}^{(t)}(\theta)=\max_{j<t} m^{(j)}(\theta)$, which attenuates updates to previously important parameters while allowing adaptation to new tasks.\\
\noindent
\textbf{Orthogonal Low-Rank Adaptation (O-LoRA)}~\cite{wang2023orthogonal} assigns each task its own low-rank adapter and constrains new updates to be orthogonal to previously learned subspaces. For each task $t$, O-LoRA introduces a new adapter $(A_t,B_t)$ while freezing prior adapters $\{(A_i,B_i)\}_{i<t}$. Let $A_t$ denote the current LoRA projection matrix and $\{A_i\}_{i<t}$ those of earlier tasks. The overlap between $A_i$ and $A_t$ subspaces is penalized using the orthogonality loss $\mathcal{L}_{\mathrm{ortho}} = \sum_{i<t} \lVert A_i A_t^\top \rVert_F^2$. The objective for task $t$ becomes $\mathcal{L} = \mathcal{L}_{\mathrm{task}} + \lambda_{\mathrm{ortho}} \mathcal{L}_{\mathrm{ortho}}$, where $\lambda_{\mathrm{ortho}}$ controls the strength of the constraint. During training, the forward pass composes all previously learned (frozen) adapters with the current adapter, while gradients update only the current one. At inference time, adapters are composed to aggregate task knowledge without modifying the frozen backbone. 

\section{Experimental Setup}
\subsection{Dataset and Evaluation Metrics}
\begin{figure}[t]
    \centering
    \includegraphics[width=\textwidth]{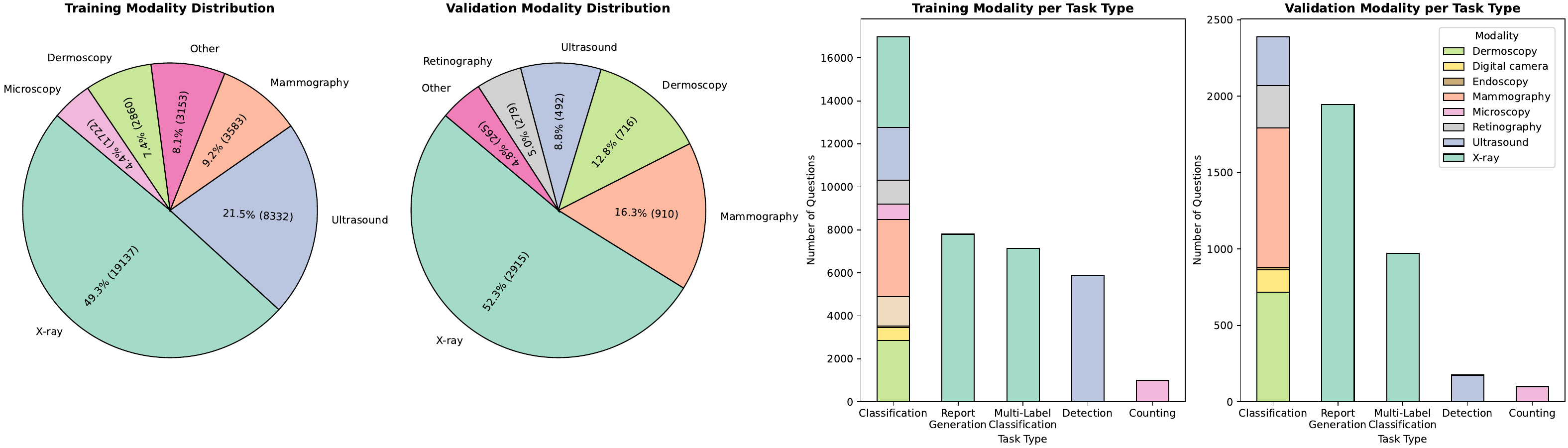} 
    \caption{Distribution of imaging modalities for training and validation sets.}
    \label{fig:dataset}
\end{figure}

We conduct experiments on the FLARE-MLLM-2D benchmark \cite{flare}, which contains image-question-answer triplets collected from 19 publicly available medical datasets. The benchmark covers eight imaging modalities and five clinical tasks, providing a comprehensive and well-suited testbed for evaluating CL under realistic task diversity. The modality distribution across splits is shown in Fig.~\ref{fig:dataset}.
Each task is assessed with a task-specific metric. Classification uses \emph{Balanced Accuracy} to account for class imbalance. Multi-label classification is evaluated with the \emph{Micro-averaged F1 Score}. Detection is measured by \emph{F1 Score} with an IoU threshold of 0.5. Cell counting uses \emph{Mean Absolute Error (MAE)}. Report generation is evaluated with the \emph{GREEN score}, which measures factual correctness and clinical relevance by capturing lexical overlap, semantic consistency, and clinical accuracy~\cite{Ostmeier_2024}. To quantify forgetting, we report the \emph{Forgetting Ratio (FR)}, defined as the relative performance drop for each task between its best performance achieved at any point during sequential training and its performance after the model has finished training on all tasks. For tasks where higher metric values are better, $\mathrm{FR}(t)=\frac{\max_{t \le i \le T} (p_t^{(i)})-p_t^{(T)}}{\max_{t \le i \le T} (p_t^{(i)})}$. For tasks where lower metric values are better, $\mathrm{FR}(t)=\frac{p_t^{(T)}-\min_{t \le i \le T} (p_t^{(i)})}{\min_{t \le i \le T} (p_t^{(i)})}$.

\subsection{Implementation Details}
We use \texttt{MedGemma-4b-it} \cite{sellergren2025medgemma} as it is pre-trained and instruction-tuned on diverse clinical tasks and imaging modalities, making it well-suited for comprehensive MedVQA evaluation. The model is loaded in 4-bit NF4 quantized format with bfloat16 precision. We apply LoRA to all linear layers with rank $r=16$, scaling factor $\alpha=32$, and dropout $0.05$. Optimization uses AdamW with a learning rate of $2\times10^{-4}$ and linear scheduling. Each experience is trained for 3 epochs with a batch size of 4 and gradient accumulation of 4 steps. The hyperparameter selection is based on the best-performing model reported on the FLARE-MLLM-2D benchmark \cite{flare,shaaban2025me}. We use a fixed replay buffer of 100 samples per task, capped by the number of available training samples for low-resource tasks. Following prior literature \cite{azimi2024kd,kirkpatrick2017overcoming,wang2023orthogonal}, we adopt standard hyperparameter settings for the rest of the CL methods: $\lambda_\mathrm{EWC}=5000$, $\lambda_{\mathrm{KD}}=0.5$, $\tau=2$, and $\lambda_{\mathrm{ortho}}=0.5$. We evaluate three curriculum strategies: 
\textbf{Order A} (Easy $\rightarrow$ Hard): 
classification, multi-label classification, detection, cell counting, report generation; 
\textbf{Order B} (Hard $\rightarrow$ Easy): 
report generation, cell counting, detection, multi-label classification, classification; 
and \textbf{Order C} (Mixed): 
classification, multi-label classification, report generation, detection, cell counting.

\section{Results and Discussion}
\begin{table*}[t!]
\centering
\caption{Performance comparison of non-CL and CL methods.
$\uparrow$ denotes higher is better, while $\downarrow$ denotes lower is better. $\mathrm{FR} \in [0,1]$ denotes the forgetting ratio (0 = no forgetting, 1 = complete forgetting). FR is not applicable for MTL and for the last task in each order, and is marked as ``--''. Each metric reflects average task performance across all sequential training stages.}
\label{tab:continual_results}
\resizebox{\textwidth}{!}{%
\renewcommand{\arraystretch}{1.15}
\begin{tabular}{
lc
>{\columncolor{clscolor!60}}c
>{\columncolor{clscolor!60}}c
>{\columncolor{mlcolor!60}}c
>{\columncolor{mlcolor!60}}c
>{\columncolor{detcolor!60}}c
>{\columncolor{detcolor!60}}c
>{\columncolor{cntcolor!60}}c
>{\columncolor{cntcolor!60}}c
>{\columncolor{rgcolor!60}}c
>{\columncolor{rgcolor!60}}c
}
\toprule
\multirow{4}{*}{Method} & \multirow{4}{*}{Order}
& \multicolumn{2}{>{\columncolor{clscolor!60}}c}{Classification}
& \multicolumn{2}{>{\columncolor{mlcolor!60}}c}{Multi-label Classification}
& \multicolumn{2}{>{\columncolor{detcolor!60}}c}{Detection}
& \multicolumn{2}{>{\columncolor{cntcolor!60}}c}{Cell Counting}
& \multicolumn{2}{>{\columncolor{rgcolor!60}}c}{Report Generation} \\

\cmidrule(lr){3-4}
\cmidrule(lr){5-6}
\cmidrule(lr){7-8}
\cmidrule(lr){9-10}
\cmidrule(lr){11-12}

& 
& Balanced Acc.~$\uparrow$ & FR~$\downarrow$
& F1 Score~$\uparrow$ & FR~$\downarrow$
& F1 Score~$\uparrow$ & FR~$\downarrow$
& MAE~$\downarrow$ & FR~$\downarrow$
& GREEN Score~$\uparrow$ & FR~$\downarrow$ \\

\midrule
MTL (Qwen2.5-VL-7B) \cite{shaaban2025me} & -- & 0.36 & -- & 0.36 & -- & 0.64 & -- & 243.60 & -- & 0.76 & -- \\
MTL (InternVL3-8B) \cite{shaaban2025me} & -- & 0.52 & -- & 0.46 & -- & 0.37 & -- & 301.43 & -- & 0.75 & -- \\
MTL (MedGemma-4B) & -- & 0.61 & -- & 0.55 & -- & 0.16 & -- & 261.26 & -- & 0.72 & -- \\
\midrule
SeqFT & A & 0.16 & 0.94 & 0.17 & 0.88 & 0.04 & 1.00 & 371.96 & 0.34 & 0.72 & -- \\

\midrule

\multirow{3}{*}{Replay}
& A 
& 0.35 & 0.50 
& 0.42 & 0.35 
& 0.05 & 1.00 
& 360.84 & 0.21 
& 0.73 & -- \\

& B 
& 0.57 & -- 
& 0.46 & 0.38 
& 0.04 & 0.95 
& 361.56 & 0.08 
& 0.62 & 0.19 \\

& C 
& 0.35 & 0.47 
& 0.41 & 0.36 
& 0.08 & 1.00 
& 323.12 & -- 
& 0.76 & 0.00 \\

\midrule

\multirow{3}{*}{EWC}
& A 
& 0.48 & 0.40
& 0.45 & 0.47
& 0.13 & 0.24
& 317.72 & 0.00
& 0.72 & -- \\

& B 
& 0.52 & --
& 0.40 & 0.34
& 0.06 & 0.83
& 263.43 & 0.02
& 0.71 & 0.01 \\

& C 
& 0.46 & 0.20
& 0.38 & 0.35
& 0.07 & 0.67
& 281.99 & --
& 0.73 & 0.00 \\

\midrule

\multirow{3}{*}{KD}
& A 
& 0.55 & 0.30 
& 0.41 & 0.38 
& 0.00 & 0.00 
& 361.88 & 0.00 
& 0.70 & -- \\

& B 
& 0.43 & -- 
& 0.39 & 0.00 
& 0.00 & 1.00 
& 389.42 & 0.10 
& 0.64 & 0.33 \\

& C 
& 0.48 & 0.20 
& 0.36 & 0.25 
& 0.00 & 0.00 
& 320.73 & --
& 0.65 & 0.13 \\

\midrule

\multirow{3}{*}{SPG}
& A 
& 0.39 & 0.85 
& 0.41 & 0.90 
& 0.03 & 1.00 
& 313.67 & 0.07 
& 0.71 & -- \\

& B 
& 0.49 & -- 
& 0.51 & 0.09 
& 0.08 & 0.68 
& 323.96 & 0.00 
& 0.34 & 0.90 \\

& C 
& 0.42 & 0.31 
& 0.34 & 0.29 
& 0.09 & 0.64 
& 232.92 & -- 
& 0.71 & 0.00 \\

\midrule

\multirow{3}{*}{O-LoRA}
& A 
& 0.38 & 1.00 
& 0.34 & 1.00 
& 0.03 & 1.00 
& 403.79 & 0.27 
& 0.71 & -- \\

& B 
& 0.44 & -- 
& 0.31 & 0.82 
& 0.03 & 1.00 
& 388.45 & 0.08 
& 0.30 & 0.99 \\

& C 
& 0.25 & 0.95 
& 0.23 & 1.00 
& 0.03 & 0.57 
& 408.00 & -- 
& 0.26 & 1.00 \\

\bottomrule
\end{tabular}}
\end{table*}

We conduct a systematic evaluation to assess the performance of existing CL methods when a medical VLM is trained sequentially on heterogeneous MedVQA tasks. Our analysis explores three key aspects: (i) the extent of catastrophic forgetting and the stability-plasticity trade-off, (ii) the sensitivity to task ordering, and (iii) the weight drift behavior of LoRA adapters. We organize our experimental analysis around three RQs. 

\noindent\textbf{RQ1. Do current CL methods mitigate catastrophic forgetting in heterogeneous MedVQA?} Results in Table~\ref{tab:continual_results} show that current CL methods mitigate but do not eliminate catastrophic forgetting under heterogeneous multimodal supervision. SEQ-FT leads to severe degradation across all tasks, confirming strong cross-task interference. Among CL approaches, EWC demonstrates the most consistent stability-plasticity balance across task types and orders. While KD occasionally achieves comparable or lower forgetting on specific tasks, it shows reduced plasticity on newly introduced objectives, such as counting and report generation, in certain orders. Replay improves over naïve fine-tuning but does not consistently control forgetting. SPG shows order-dependent instability, and O-LoRA exhibits larger forgetting spikes. Performance in detection remains consistently low across all configurations. This trend is consistent with prior studies, which report limited detection performance even when varying backbone architectures or fine-tuning task-specific LoRA \cite{flare,shaaban2025me}.

\begin{figure}[t]
    \centering
    \includegraphics[width=\textwidth]{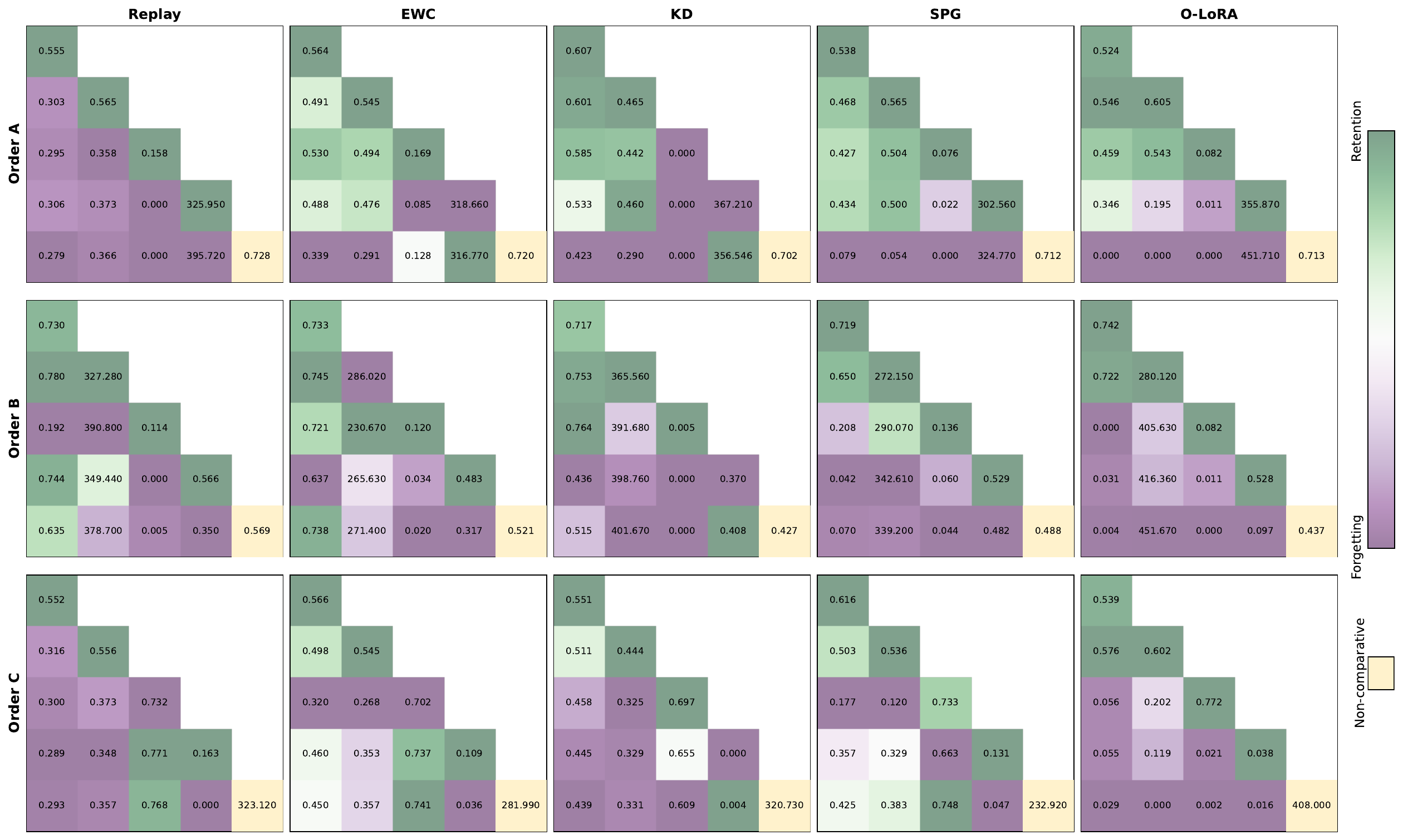} 
    \caption{Performance and forgetting trends across CL methods. Colors encode relative performance change across training stages.}
    \label{fig:FR}
\end{figure}

\noindent\textbf{RQ2. How does task ordering influence performance and forgetting across different CL methods?} The triangular matrices in Fig.~\ref{fig:FR}, where each entry($i,j$) represents performance on Task$_j$ after training on Task$_i$, show that task ordering significantly affects both final performance and forgetting behavior, and sensitivity varies across methods. KD and EWC are comparatively more stable to permutation, exhibiting smaller performance fluctuations across curricula. O-LoRA is highly sensitive to ordering, while Replay and SPG show intermediate sensitivity. We observed that a central factor driving order effects is report generation. As a long-form autoregressive objective, it shifts the model toward producing extended free-form text rather than concise, format-constrained outputs. After training on report generation, most methods generate longer responses even for tasks requiring short answers, altering previously learned output formats, and increasing forgetting when introduced early. This behavior is less evident in KD, as output distillation constrains decoding shifts.

\begin{figure}[htbp!]
    \centering
    \includegraphics[width=\textwidth]{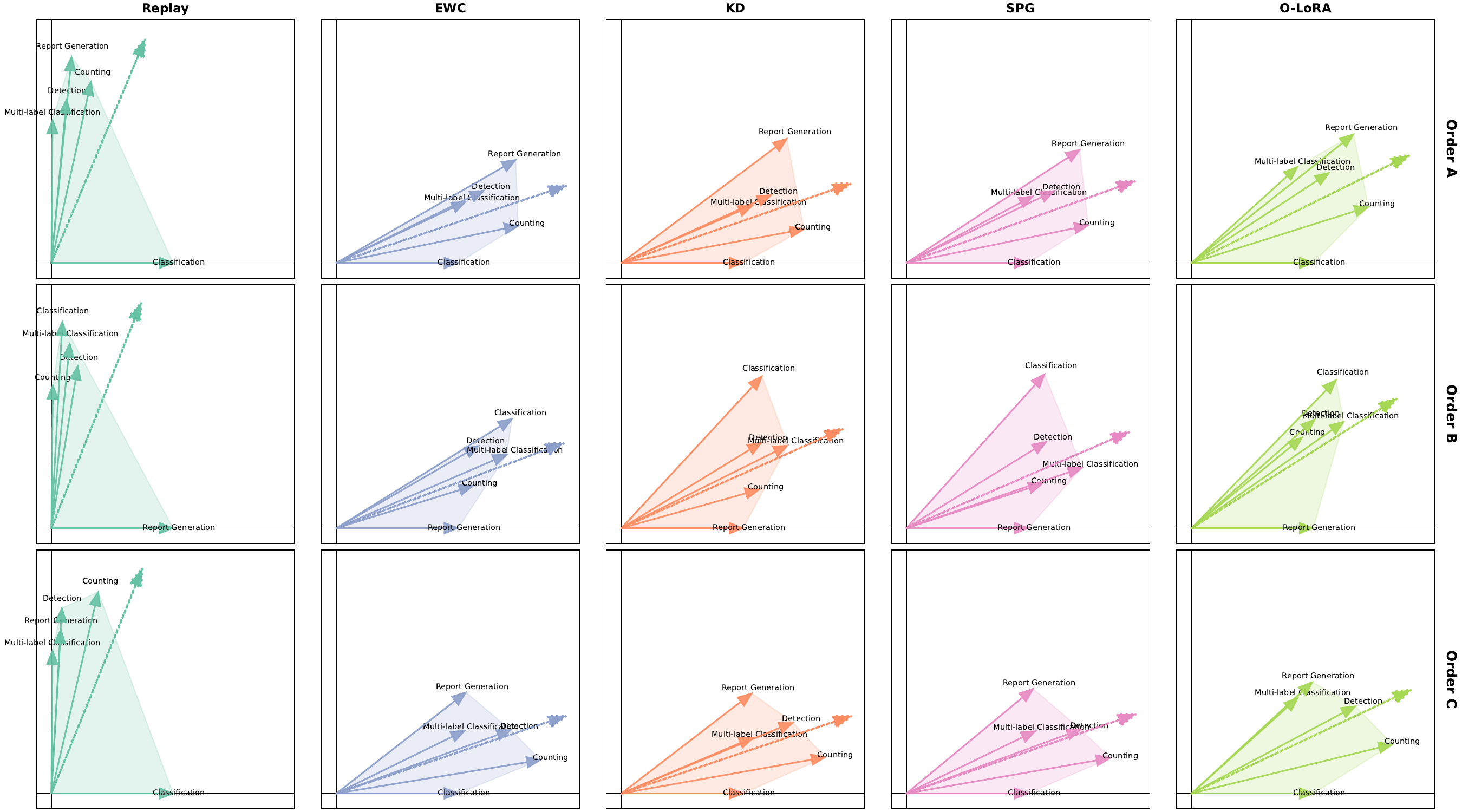} 
    \caption{LoRA parameter directions across sequential tasks. Angles are computed in the high-dimensional LoRA space using Frobenius inner-product similarity between $t-1$ and $t$ and visualized as 2D arrows via polar mapping. The dotted arrow shows the mean drift direction.}
    \label{fig:lora}
\end{figure}
\noindent\textbf{RQ3. How does drift in LoRA weights relate to forgetting behavior across CL methods?} In Fig.~\ref{fig:lora}, we visualize the angles between the LoRA directions, measuring similarity via the Frobenius inner product $\mathrm{sim}(\mathbf{t_1}, \mathbf{t_2}) = \frac{\langle \mathbf{t_1}, \mathbf{t_2} \rangle_F}{\|\mathbf{t_1}\|_F \, \|\mathbf{t_2}\|_F}$, where $\langle \mathbf{t_1}, \mathbf{t_2} \rangle_F = \mathrm{Tr}(\mathbf{t}_1 \mathbf{t}_2^{\top})$. The figure reveals distinct LoRA parameter shift patterns across CL methods that align with their observed forgetting behavior. Replay exhibits a broad angular spread, indicating substantial directional drift as new tasks are introduced. In contrast, EWC and KD maintain tighter distributions, suggesting closer alignment with earlier tasks, thus improving stability. SPG shows moderate dispersion. O-LoRA exhibits progressive directional shift across tasks, indicating cumulative drift. When report generation is introduced early (Orders B and C), the initial angular deviation increases, and the overall drift is amplified. Overall, these results indicate that CL methods differ in reshaping the underlying low-rank parameter space in ways that correlate with the observed forgetting behavior.

\section{Conclusion}
This study presents a comprehensive evaluation of CL for heterogeneous MedVQA, including classification, multi-label classification, detection, cell counting, and report generation. Our results show that existing CL strategies struggle to mitigate catastrophic forgetting and maintain a stability-plasticity balance when tasks differ in objectives and supervision formats. Performance retention varies significantly across methods and is highly sensitive to task order, indicating limited robustness in realistic sequential training. Analysis of LoRA dynamics further shows that some methods maintain closer alignment of parameter updates across tasks, whereas others induce larger directional shifts, reflecting varying degrees of cross-task interference. These findings show that parameter-level constraints alone are insufficient; robust CL for heterogeneous MedVQA also requires explicit functional alignment to preserve output behavior across arbitrary task sequences. Future work may investigate more advanced methods within the most promising CL strategies, as well as evaluate larger VLMs and 3D medical imaging tasks.

%
%
%
\bibliographystyle{splncs04}
\bibliography{refs}

@article{lwf,
	title        = {Learning without forgetting},
	author       = {Li, Zhizhong and Hoiem, Derek},
	year         = 2017,
	journal      = {IEEE transactions on pattern analysis and machine intelligence},
	publisher    = {IEEE},
	volume       = 40,
	number       = 12,
	pages        = {2935--2947}
}

@inproceedings{rebuffi2017icarl,
	title        = {icarl: Incremental classifier and representation learning},
	author       = {Rebuffi, Sylvestre-Alvise and Kolesnikov, Alexander and Sperl, Georg and others},
	year         = 2017,
	booktitle    = {Proceedings of the IEEE conference on Computer Vision and Pattern Recognition},
	pages        = {2001--2010}
}

@inproceedings{wang2023orthogonal,
	title        = {Orthogonal subspace learning for language model continual learning},
	author       = {Wang, Xiao and Chen, Tianze and Ge, Qiming and others},
	year         = 2023,
	booktitle    = {Findings of the Association for Computational Linguistics: EMNLP 2023},
	pages        = {10658--10671}
}

@inproceedings{konishi2023parameter,
	title        = {Parameter-Level Soft-Masking for Continual Learning},
	author       = {Konishi, Tatsuya and Kurokawa, Mori and Ono, Chihiro and others},
	year         = 2023,
	booktitle    = {Proceedings of the 40th International Conference on Machine Learning}
}

@article{kirkpatrick2017overcoming,
	title        = {Overcoming catastrophic forgetting in neural networks},
	author       = {Kirkpatrick, James and Pascanu, Razvan and Rabinowitz, Neil and others},
	year         = 2017,
	journal      = {Proceedings of the National Academy of Sciences},
	volume       = 114,
	number       = 13,
	pages        = {3521--3526},
	doi          = {10.1073/pnas.1611835114}
}

@article{azimi2024kd,
	title        = {KD-LoRA: A Hybrid Approach to Efficient Fine-Tuning with LoRA and Knowledge Distillation},
	author       = {Azimi, Rambod and Rishav, Rishav and Teichmann, Marek and others},
	year         = 2024,
	journal      = {arXiv preprint arXiv:2410.20777}
}

@inproceedings{bagus2021investigation,
	title        = {An Investigation of Replay-based Approaches for Continual Learning},
	author       = {Bagus, Benedikt and Gepperth, Alexander},
	year         = 2021,
	booktitle    = {2021 International Joint Conference on Neural Networks (IJCNN)},
	publisher    = {IEEE},
	pages        = {1--9}
}

@inproceedings{merlin2022practical,
	title        = {Practical Recommendations for Replay-Based Continual Learning Methods},
	author       = {Merlin, Gabriele and Lomonaco, Vincenzo and Cossu, Andrea and others},
	year         = 2022,
	booktitle    = {Image Analysis and Processing -- ICIAP 2022 Workshops},
	publisher    = {Springer},
	pages        = {548--559}
}

@article{lin2023medical,
	title        = {Medical visual question answering: A survey},
	author       = {Lin, Zhihong and Zhang, Donghao and Tao, Qingyi and others},
	year         = 2023,
	journal      = {Artificial Intelligence in Medicine},
	publisher    = {Elsevier},
	volume       = 143,
	pages        = 102611
}

@article{kalpelbe2025vision,
	title        = {Vision language models in medicine},
	author       = {Kalp{\'e}lb{\'e}, B{\'e}ria Chingnab{\'e} and Adaambiik, Angel Gabriel and Peng, Wei},
	year         = 2025,
	journal      = {arXiv preprint arXiv:2503.01863}
}

@article{bruno2025continual,
	title        = {Continual learning in medicine: A systematic literature review},
	author       = {Bruno, Pierangela and Quarta, Alessandro and Calimeri, Francesco},
	year         = 2025,
	journal      = {Neural Processing Letters},
	publisher    = {Springer},
	volume       = 57,
	number       = 1,
	pages        = 2
}

@article{qazi2024continual,
	title        = {Continual learning in medical imaging: a survey and practical analysis},
	author       = {Qazi, Mohammad Areeb and Hashmi, Anees Ur Rehman and Sanjeev, Santosh and others},
	year         = 2024,
	journal      = {ACM Computing Surveys},
	publisher    = {ACM New York, NY}
}

@article{wang2024comprehensive,
	title        = {A comprehensive survey of continual learning: Theory, method and application},
	author       = {Wang, Liyuan and Zhang, Xingxing and Su, Hang and others},
	year         = 2024,
	journal      = {IEEE transactions on pattern analysis and machine intelligence},
	publisher    = {IEEE},
	volume       = 46,
	number       = 8,
	pages        = {5362--5383}
}

@book{chen2018lifelong,
	title        = {Lifelong machine learning},
	author       = {Chen, Zhiyuan and Liu, Bing},
	year         = 2018,
	publisher    = {Morgan \& Claypool Publishers}
}

@article{qu2025recent,
	title        = {Recent advances of continual learning in computer vision: An overview},
	author       = {Qu, Haoxuan and Rahmani, Hossein and Xu, Li and others},
	year         = 2025,
	journal      = {IET Computer Vision},
	publisher    = {Wiley Online Library},
	volume       = 19,
	number       = 1,
	pages        = {e70013}
}

@inproceedings{biesialska2020continual,
	title        = {Continual lifelong learning in natural language processing: A survey},
	author       = {Biesialska, Magdalena and Biesialska, Katarzyna and Costa-Jussa, Marta R},
	year         = 2020,
	booktitle    = {Proceedings of the 28th international conference on computational linguistics},
	pages        = {6523--6541}
}

@article{hu2022lora,
	title        = {Lora: Low-rank adaptation of large language models.},
	author       = {Hu, Edward J and Shen, Yelong and Wallis, Phillip and others},
	year         = 2022,
	journal      = {Iclr},
	volume       = 1,
	number       = 2,
	pages        = 3
}

@misc{flare,
	title        = {{FLARE-Task5-MLLM-2D} Medical Multimodal Dataset},
	author       = {{FLARE-MedFM}},
	year         = 2025,
	note         = {Accessed: 2026},
	howpublished = {\url{https://huggingface.co/datasets/FLARE-MedFM/FLARE-Task5-MLLM-2D}},
	license      = {CC-BY-NC-4.0}
}

@inproceedings{Ostmeier_2024,
	title        = {GREEN: Generative Radiology Report Evaluation and Error Notation},
	author       = {Ostmeier, Sophie and Xu, Justin and Chen, Zhihong and others},
	year         = 2024,
	booktitle    = {Findings of the Association for Computational Linguistics: EMNLP 2024},
	publisher    = {Association for Computational Linguistics},
	pages        = {374–390},
	doi          = {10.18653/v1/2024.findings-emnlp.21},
	url          = {http://dx.doi.org/10.18653/v1/2024.findings-emnlp.21}
}

@inproceedings{jiang2025omnidoctor,
	title        = {OmniDoctor: Towards LLM-centric Lifelong Learning for New Emerging Medical VQA Tasks},
	author       = {Jiang, Na and Zheng, Wenhui and Gu, Xuqian and others},
	year         = 2025,
	booktitle    = {Proceedings of the 33rd ACM International Conference on Multimedia},
	pages        = {6567--6575}
}

@inproceedings{shaaban2025motor,
	title        = {MOTOR: Multimodal Optimal Transport via Grounded Retrieval in Medical Visual Question Answering},
	author       = {Shaaban, Mai A and Saleem, Tausifa Jan and Papineni, Vijay Ram Kumar and others},
	year         = 2025,
	booktitle    = {International Conference on Medical Image Computing and Computer-Assisted Intervention},
	pages        = {459--469},
	organization = {Springer}
}

@article{sellergren2025medgemma,
	title        = {MedGemma Technical Report},
	author       = {Sellergren, Andrew and Kazemzadeh, Sahar and Jaroensri, Tiam and others},
	year         = 2025,
	journal      = {arXiv preprint arXiv:2507.05201}
}

@article{peft,
  title={Parameter-efficient fine-tuning for large models: A comprehensive survey},
  author={Han, Zeyu and Gao, Chao and Liu, Jinyang and Zhang, Jeff and Zhang, Sai Qian},
  journal={arXiv preprint arXiv:2403.14608},
  year={2024}
}

@inproceedings{shaaban2025me,
  title={ME-VLIP: A Modular and Efficient Vision-Language Framework for Generalizable Medical Image Parsing},
  author={Shaaban, Mai A and Saqib, Amal and Hardan, Shahad Emad and Taratynova, Darya and Saleem, Tausifa Jan and Yaqub, Mohammad},
  booktitle={MICCAI 2025 FLARE Challenge},
  year=2025
}

@article{lau2018dataset,
  title={A dataset of clinically generated visual questions and answers about radiology images},
  author={Lau, Jason J and Gayen, Soumya and Ben Abacha, Asma and Demner-Fushman, Dina},
  journal={Scientific data},
  volume={5},
  number={1},
  pages={180251},
  year={2018},
  publisher={Nature Publishing Group}
}

@incollection{chen2022continual,
  title={Continual learning and catastrophic forgetting},
  author={Chen, Zhiyuan and Liu, Bing},
  booktitle={Lifelong Machine Learning},
  pages={55--75},
  year={2022},
  publisher={Springer}
}

\end{document}